\documentclass{MITcsail}

\usepackage{amsmath,amsfonts,bm}

\def\eqref#1{equation~\ref{#1}}

\def\1{\bm{1}}

\DeclareMathAlphabet{\mathsfit}{\encodingdefault}{\sfdefault}{m}{sl}
\SetMathAlphabet{\mathsfit}{bold}{\encodingdefault}{\sfdefault}{bx}{n}

\usepackage{amsmath}
\usepackage{amsfonts}
\usepackage{amssymb}
\usepackage{wrapfig}
\usepackage{subcaption}
\usepackage{multirow}
 \usepackage{mathtools} 

\usepackage{verbatim}

\usepackage{anyfontsize}

\usepackage{microtype}
\usepackage{graphicx}
\usepackage{booktabs} 
\usepackage{multirow}
\usepackage{amsmath,amssymb}
\usepackage{booktabs}
\usepackage{caption,subcaption}

\usepackage{xcolor}
\definecolor{mygreen}{HTML}{3cb44b}
\definecolor{skyblue}{HTML}{beffff}
\definecolor{lightgreen}{HTML}{90ee90}

\usepackage{color, colortbl}

\definecolor{emerald}{rgb}{0.31, 0.78, 0.37}

\usepackage{tcolorbox}
\usepackage{enumitem}
\usepackage{colortbl}

\usepackage{xcolor}
\definecolor{mygreen}{HTML}{3cb44b}
\colorlet{myyellow}{green!10!orange!90!}
\makeatletter

\usepackage{tikz}
\usetikzlibrary{arrows,shapes,snakes,automata,backgrounds,fit,petri}
\usepackage{adjustbox}

\newcommand{\RN}[1]{%
	\textup{\lowercase\expandafter{\it \romannumeral#1}}%
}
\usepackage{tabu}

\newcommand{\beq}{\vspace{0mm}\begin{equation}}
\newcommand{\eeq}{\vspace{0mm}\end{equation}}
\newcommand{\beqs}{\vspace{0mm}\begin{eqnarray}}
\newcommand{\eeqs}{\vspace{0mm}\end{eqnarray}}
\newcommand{\barr}{\begin{array}}
\newcommand{\earr}{\end{array}}

\usepackage{color, colortbl}
\definecolor{Gray}{gray}{0.93}

\usepackage{lipsum}

\usepackage{pifont}

\usepackage{makecell}

\usepackage{xcolor,amsmath}
\usepackage[linesnumbered,ruled,vlined]{algorithm2e}
\DontPrintSemicolon

\usepackage{xcolor}
\definecolor{mygreen}{HTML}{3cb44b}

\SetKwComment{Comment}{\color{green!50!black}\# }{}

\SetKwProg{Function}{def}{:}{}

\SetKwProg{For}{for}{:}{}
\SetKwProg{If}{if}{:}{}

\usepackage{hyperref}
\usepackage{xcolor}
\hypersetup{
    colorlinks=true,
    linkcolor=orange,
    citecolor=lightgray,
    filecolor=darkred,
    urlcolor=orange
}
\usepackage[numbers, sort&compress]{natbib}
\usepackage{float}
\usepackage{subcaption}
\usepackage{bbm}

\definecolor{darkred}{RGB}{140, 21, 21}
\definecolor{lightgray}{gray}{0.7}
\definecolor{orange}{HTML}{FF671F}

\title{Datasets and Recipes for Video Temporal Grounding via Reinforcement Learning}

\author{Ruizhe Chen~$^{1,2}$\footnote{Work done during his internship at Bytedance.}, Zhiting Fan~$^{2}$, Tianze Luo~$^{1}$, Heqing Zou~$^{1}$, Zhaopeng Feng~$^{2}$, 
\\
Guiyang Xie~$^{1}$, Hansheng Zhang~$^{1}$, Zhuochen Wang~$^{1}$, Zuozhu Liu~$^{2}$, Huaijian Zhang~$^{1}$
\\
\vspace{1em}
\normalfont{\small $^{1}$ Bytedance}\\
\normalfont{\small $^{2}$ Zhejiang University}\\
\vspace{1em}
\texttt{Link: 
\href{https://github.com/zjuruizhechen/TVG-R1}{Code \& Model}}
\vspace{1em}
}

\begin{document}

\maketitle
\thispagestyle{firstpagestyle} 

\begin{abstract}
Video Temporal Grounding (VTG) aims to localize relevant temporal segments in videos given natural language queries. Despite recent progress with large vision-language models (LVLMs) and instruction-tuning, existing approaches often suffer from limited temporal awareness and poor generalization.
In this work, we introduce a two-stage training framework that integrates supervised fine-tuning with reinforcement learning (RL) to improve both the accuracy and robustness of VTG models. Our approach first leverages high-quality curated cold start data for SFT initialization, followed by difficulty-controlled RL to further enhance temporal localization and reasoning abilities. Comprehensive experiments on multiple VTG benchmarks demonstrate that our method consistently outperforms existing models, particularly in challenging and open-domain scenarios.
We conduct an in-depth analysis of training strategies and dataset curation, highlighting the importance of both high-quality cold start data and difficulty-controlled RL. To facilitate further research and industrial adoption, we release all intermediate datasets, models, and code to the community.

\end{abstract}

\section{Introduction}

With the proliferation of social media platforms, video content has become the most information-rich and diverse medium for capturing and conveying daily experiences. As a result, efficiently identifying specific moments within videos based on user queries—a task known as \emph{Video Temporal Grounding} (VTG)—has emerged as a core capability for a range of industrial applications, from intelligent video retrieval to workflow optimization and automated event monitoring~\cite{Grauman2022Ego4D,Hendricks2017MCN,Li2023VideoChat}.
VTG enables practitioners to swiftly pinpoint relevant segments in massive videos, significantly reducing manual review workloads and empowering real-time decision-making~\cite{sultani2018real}. 

Recent advances in large vision-language models (LVLMs) have led to the development of end-to-end temporal grounding frameworks. Instruction-tuned models such as TimeChat~\cite{Ren2024TimeChat}, VTimeLLM~\cite{Huang2024VTimeLLM}, and LITA~\cite{Huang2024LITA} reformulate temporal grounding as a text generation task, while models like Momentor~\cite{Qian2024Momentor} and VTG-LLM~\cite{Guo2024VTG-LLM} introduce specialized modules or vocabulary to improve temporal perception. Despite notable progress, existing approaches are still constrained by the inherent limitations of supervised fine-tuning, struggling with precise temporal awareness and generalization.

To address these challenges, we propose a novel two-stage training framework that integrates supervised fine-tuning (SFT) with reinforcement learning (RL) to significantly improve the performance and generalization of open-source models for VTG tasks. Our framework first leverages high-quality curated data to provide the model with a robust \emph{coldstart} initialization via SFT, followed by a difficulty-controlled RL stage that further enhances temporal grounding abilities and reasoning.

We conduct extensive experiments across multiple VTG benchmarks, systematically evaluating the contributions of each training stage. Our findings highlight the critical importance of high-quality cold-start data and controlled RL training, providing actionable insights for practical deployment in real-world industrial scenarios. Furthermore, to facilitate future research and application, we release all intermediate results and code as open-source resources.

\begin{figure*}[htbp]
    \centering
    \includegraphics[width=1.0\textwidth]{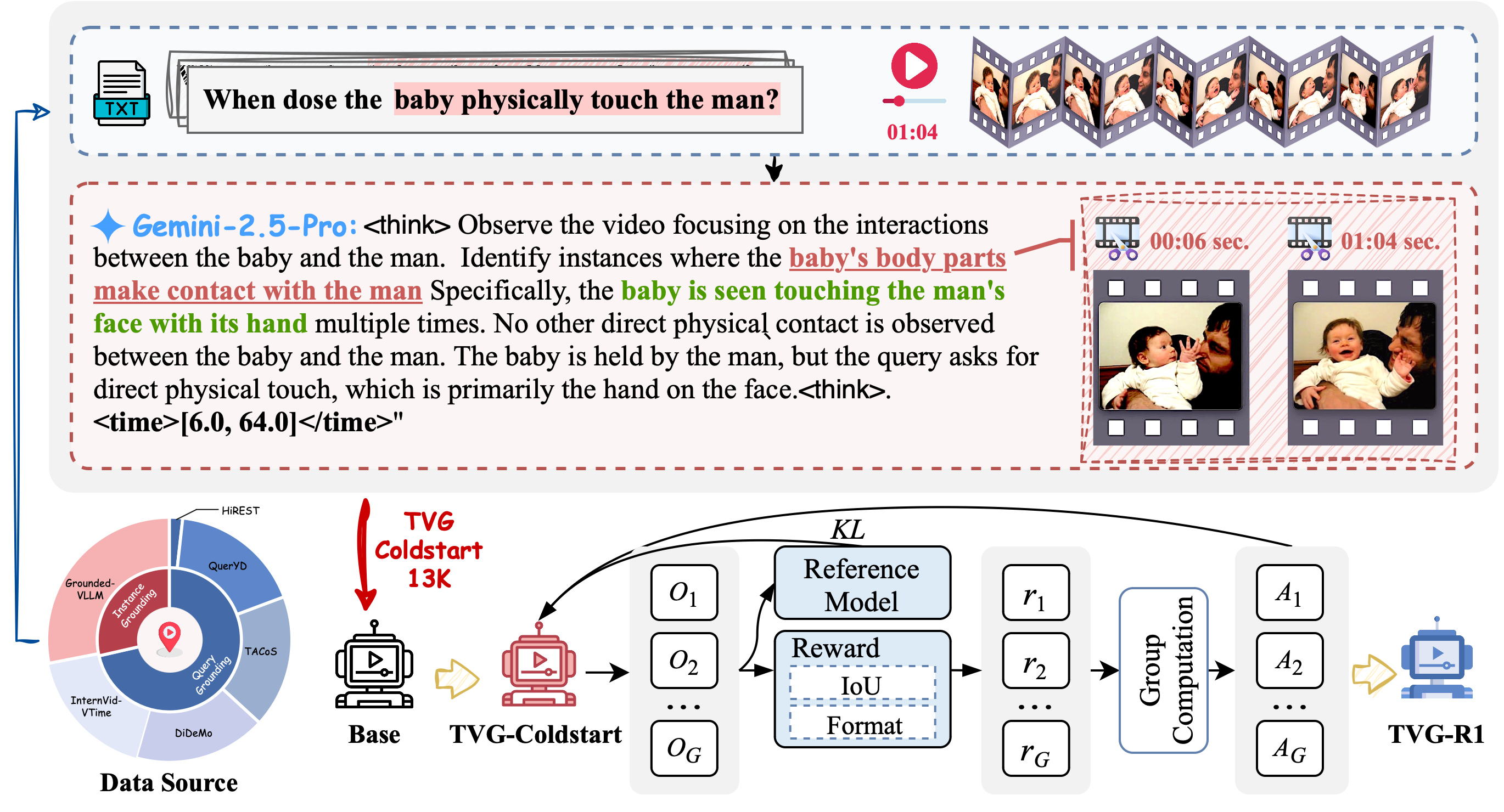}
    \caption{\textbf{Overview of the proposed training pipeline for Video Temporal Grounding (VTG-R1)}. The framework first performs supervised fine-tuning (SFT) with curated high-quality cold-start data to initialize the base model, followed by reinforcement learning (RL) to further enhance temporal localization abilities. }
    \label{fig:tvg_pipeline}
\end{figure*}

The main contributions of this work are:
\begin{itemize}
    \item We introduce a two-stage training framework that combines SFT and RL to advance open-source LVLMs for video temporal grounding.
    \item We conduct comprehensive evaluations across multiple benchmarks, validating the effectiveness and scalability of our approach.
    \item We open-source all intermediate datasets, models, and code to support further research and industrial adoption.
\end{itemize}

\section{Related Works}

Video Temporal Grounding (VTG) aims to localize relevant temporal segments within untrimmed videos given natural language queries~\cite{Grauman2022Ego4D,Hendricks2017Moment,Li2023VideoChat,Dai2023InstructBLIP,Wang2022NegativeSample}. Early efforts, such as CTRL and MCN, introduced foundational approaches that leveraged sliding windows and dual-stream networks to generate candidate segments for text-video matching~\cite{Gao2017TALL,Hendricks2017MCN}, which laid the groundwork for subsequent advancements.


With the emergence of large vision-language models (LVLMs), recent research has shifted towards end-to-end VTG frameworks that leverage instruction-tuning and textual generation. Models such as TimeChat~\cite{Ren2024TimeChat}, VTimeLLM~\cite{Huang2024VTimeLLM}, and LITA~\cite{Huang2024LITA} reformulate temporal grounding as a sequence generation task, while Momentor~\cite{Qian2024Momentor} addresses temporal quantization errors by introducing temporal-aware modules. Other approaches, including Grounded-VideoLLM and VTG-LLM~\cite{Wang2024GroundedVideoLLM,Guo2024VTG-LLM}, expand model vocabularies to facilitate the learning of temporal embeddings, further improving grounding precision.

VTG technology has shown practical value in diverse domains. In manufacturing, VTG supports automated workflow analysis and anomaly detection to improve operational efficiency~\cite{li2021vision}. For security surveillance, VTG enables fast retrieval of critical events, supporting both real-time monitoring and retrospective investigation~\cite{sultani2018real}. In healthcare, VTG facilitates efficient identification of key procedures in large-scale surgical videos, benefiting both clinical analysis and education~\cite{twinanda2017endonet}.

Despite these advances, the predominant reliance on supervised fine-tuning (SFT) often restricts the model's temporal awareness and generalization capabilities, especially in open-domain or challenging scenarios. To address these limitations, we propose a two-stage training framework that integrates supervised fine-tuning with reinforcement learning, aiming to enhance both the accuracy and generalization of VTG models. To support further research and application, we release all intermediate data, models, and code as open-source resources.



\begin{table*}[htbp]
\centering
\resizebox{1.0\textwidth}{!}{
\begin{tabular}{lclcc}
\toprule
Task & \# Original Samples & Source Datasets & \# Coldstart Samples & \# RL Samples \\
\midrule
\makecell[l]{Instance Grounding \\ (Moment Retrieval)} & 40K &
\makecell[l]{HiREST~\cite{zala2023hierarchical} (4K),\\ QuerYD~\cite{oncescu2021queryd} (33K),\\ TACoS~\cite{gan2023temporal} (10K),\\ DiDeMo~\cite{anne2017didemo} (33K),\\ InternVid-VTime~\cite{wang2023internvid} (54K)} 
& 10K & 13K  \\
\midrule
Query Grounding & 16K & Grounded-VLLM~\cite{Wang2024GroundedVideoLLM} (16K) & 3K & 5K \\
\midrule
Total & 56K & - & 13K & 18K \\
\bottomrule
\end{tabular}
}
\caption{Statistics of the source datasets and filtered coldstart and RL datasets.}
\label{tab:Overview of tasks}
\end{table*}

\section{Datasets and Recipes}

In this section, we present the detailed process for constructing VTG-R1 via a two-stage training pipeline, encompassing data collection, curation, and specific training procedures.

\subsection{Data Collection and Curation}

High-quality coldstart and RL datasets are essential for enhancing the temporal video grounding capabilities of MLLMs. Here, we describe our approach to collecting source data and curating the TVG-RL-18K dataset for RL training and the TVG-Coldstart-13K dataset for SFT-based coldstart.

\paragraph{Data Collection.}
 
We aggregate data from various public datasets, including those for moment retrieval and query grounding tasks, carefully sampling and balancing the proportion of each subset. The distributions of the raw source data for TVG-RL-18K and TVG-Coldstart-13K are categorized and summarized in Table~\ref{tab:Overview of tasks}.

\paragraph{CoT Annotation and Data Filtering.}

To enable effective supervised fine-tuning (SFT) cold-start, we employ Gemini-2.5-Pro to generate chain-of-thought (CoT) rationales for the source samples. The prompt template used for CoT generation is provided below and is consistently applied during both the SFT and RL stages. We then filter the annotated samples according to their final Intersection-over-Union (IoU) scores: samples with IoU $> \epsilon_1$ are regarded as high-quality and their CoT rationales are retained for cold-start, forming the TVG-Coldstart-13K subset. In contrast, source samples with IoU $< \epsilon_2$ are considered low-quality—often due to excessive difficulty or annotation errors—and are excluded from the RL stage. The remaining samples constitute the TVG-RL-18K subset.

\begin{tcolorbox}[
  colback=gray!10!white,
  colframe=black,
  coltitle=black,
  title=\textbf{Prompt Template for TVG-R1},
  colbacktitle=gray!30!white,
  boxrule=0.5pt,
  rounded corners,
]
\texttt{system\ } 
You MUST reason based on the temporal changes and visual evidence in the video to determine the precise time period related to the query. The reasoning MUST reflect how the content evolves over time, not general logic.
The reasoning process MUST BE enclosed within $\langle$ think$\rangle$ $\langle$/think$\rangle$ tags. The specific time period MUST BE in the format [start time, end time] in seconds enclosed within $\langle$time$\rangle$ $\langle$/time$\rangle$ tags.\\
\texttt{user\ }\texttt{\{query / instance\}} 

\end{tcolorbox}

\subsection{Supervised Fine-Tuning (SFT) Stage}

In the first stage of our training pipeline, we employ supervised fine-tuning (SFT) to provide the model with a high-quality initialization, referred to as the \textit{cold start} phase. This process equips the model with robust multimodal alignment and structured reasoning capabilities from the outset, laying a solid foundation for the subsequent reinforcement learning stage.

\subsection{Reinforcement Learning (RL) Stage}

\subsubsection{Reward Modeling}
\label{sec:Reward Modeling}

The reward \( r_i \) plays a crucial role in guiding the model's learning objective. To promote effective temporal grounding with explicit reasoning, we employ a composite reward function consisting of two components: the IoU reward \( r_{\text{tIoU}} \) and the reasoning format reward \( r_{\text{form}} \).

\paragraph{Timestamp-aware IoU Reward \( r_{\text{tIoU}}(\cdot) \)}

In the TVG task, the quality of a predicted temporal segment \([t_s, t_e]\) is primarily evaluated using the Intersection-over-Union (IoU) metric, which measures the overlap between the predicted segment and the ground-truth segment \([t'_s, t'_e]\). The IoU is computed as:
\[
r_{\text{tIoU}} = \frac{[t_s, t_e] \cap [t'_s, t'_e]}{[t_s, t_e] \cup [t'_s, t'_e]}
\]
where \(\cap\) and \(\cup\) denote the intersection and union of the predicted and ground-truth intervals.

\paragraph{Reasoning Format Reward \( r_{\text{form}}(\cdot) \)}
To explicitly encourage the model to generate responses with structured reasoning, we introduce a format-based reward \( r_{\text{form}} \), which verifies whether the output follows the expected reasoning format. Specifically, we require the model to enclose the reasoning process within \texttt{<think>...</think>} tags and the final answer within \texttt{<answer>...</answer>} tags. The reward is defined as:
\[
r_{\text{form}} = \mathbbm{1}_{\{\texttt{<think>}, \texttt{</think>}, \texttt{<answer>}, \texttt{</answer>}\} \subseteq \text{output}}
\]
where \( \mathbbm{1}_{\cdot} \) denotes the indicator function. 

\paragraph{Final Reward \( r_i \)}

The final reward \( r_i \) is defined as a weighted sum of the two components:
\[
r_i = \lambda_{\text{tIoU}} \cdot r_{\text{tIoU}} + \lambda_{\text{form}} \cdot r_{\text{form}}
\]
where \( \lambda_{\text{tIoU}} \) and \( \lambda_{\text{form}} \) are hyperparameters.

\begin{table*}[t]
\centering
\resizebox{\linewidth}{!}{
\begin{tabular}{l|cccc|cccc}
\toprule
\multirow{2}{*}{\textbf{Model}} & \multicolumn{4}{c|}{\textbf{NExTGQA}} & \multicolumn{4}{c}{\textbf{RexTime}} \\
 & \textbf{R@0.3} & \textbf{R@0.5} & \textbf{R@0.7} & \textbf{mIoU} & \textbf{R@0.3} & \textbf{R@0.5} & \textbf{R@0.7} & \textbf{mIoU} \\
\midrule
Qwen2.5-VL-7B thinking    & 25.81 & 14.73 & 8.72  & 17.74 & 12.16 & 7.17  & 2.71 & 10.17 \\
Qwen2.5-VL-7B & 31.60 & 18.06 & 7.46 & 20.87 & 10.31 & 6.08 & 3.04 & 8.10 \\
Qwen2.5-VL-32B & 37.96 & 22.26 & 9.98 & 25.35 & 16.83 & 9.99 & 5.10 & 13.02 \\
VTimeLLM                  & 37.90 & 20.20 & 9.71  & 24.40 & 28.84 & 17.41 & 7.19 & 20.14 \\
TimeChat                  & 34.10 & 17.90 & 6.24  & 20.60 & 14.42 & 7.61  & 3.06 & 11.65 \\
VideoChat-TPO             & 41.20 & \textbf{23.40} & 8.15  & 27.70 & 34.53 & 19.26 & 9.81 & 25.23 \\
\midrule
TVG-ColdStart             & 21.74 & 11.54 & 5.24 & 15.09 & 13.57 & 7.82 & 4.34 & 10.18    \\
TVG-R1                    & \textbf{41.65} & 20.78 & \textbf{10.01} & \textbf{29.25} & \textbf{41.04} & \textbf{24.54} & \textbf{11.07}& \textbf{28.20} \\
\bottomrule
\end{tabular}
}
\caption{Performance comparison on NExTGQA and RexTime benchmarks. It can be observed that VTG-R1 outperforms existing SFT-based methods trained with large-scale data.}
\label{tab:main_results}
\end{table*}

\subsubsection{GRPO Training}
We adopt Group Relative Policy Optimization (GRPO)~\cite{shao2024deepseekmath} for reinforcement learning, which is a variant of Proximal Policy Optimization (PPO)~\cite{schulman2017ppo}. Unlike PPO, which relies on a learned critic, GRPO directly compares a group of candidate responses, removing the need for a critic model and thereby reducing computational overhead.

Given a query \( q \), GRPO samples \( G \) distinct candidate responses \( o = \{o_1, \dots, o_G\} \) from the policy. Rewards for each response are assigned as described in Sec.~\ref{sec:Reward Modeling}, yielding \( \{r_1, \dots, r_G\} \). These scores are then normalized within the group, and the advantage of each response is defined as:
\begin{align}
A_i = \frac{r_i - \mu}{\sigma}, \quad \text{where  } \nonumber
\mu = \frac{1}{G} \sum_{j=1}^{G} r_j, 
\sigma = \sqrt{\frac{1}{G} \sum_{j=1}^{G} (r_j - \mu)^2}.\nonumber
\label{eq:grpo_advantage}
\end{align}
Here, \( A_i \) denotes the normalized advantage of the \( i \)-th response. GRPO encourages the model to assign higher probabilities to relatively better responses within the group.
The final training objective also includes a KL-divergence regularization term to prevent the updated policy \( \pi_\theta \) from deviating significantly from a reference policy \( \pi_{\text{ref}} \). The complete objective is given by:
\begin{align}
\mathcal{L}_{\text{GRPO}} = \mathbb{E}_{o \sim \pi_{\theta_{\text{old}}}(p)} & \Bigg[ 
 \sum_{i=1}^{G} \frac{\pi_\theta(o_i)}{\pi_{\theta_{\text{old}}}(o_i)} \cdot A_i \notag
 - \beta \cdot D_{\text{KL}} \left( \pi_\theta \,\|\, \pi_{\text{ref}} \right) 
\Bigg], \nonumber
\end{align}
where \( \beta \) is a regularization coefficient controlling the divergence from the reference policy.

\begin{figure}[htbp]
    \centering
    \begin{subfigure}[t]{0.31\textwidth}
        \centering
        \includegraphics[width=\textwidth]{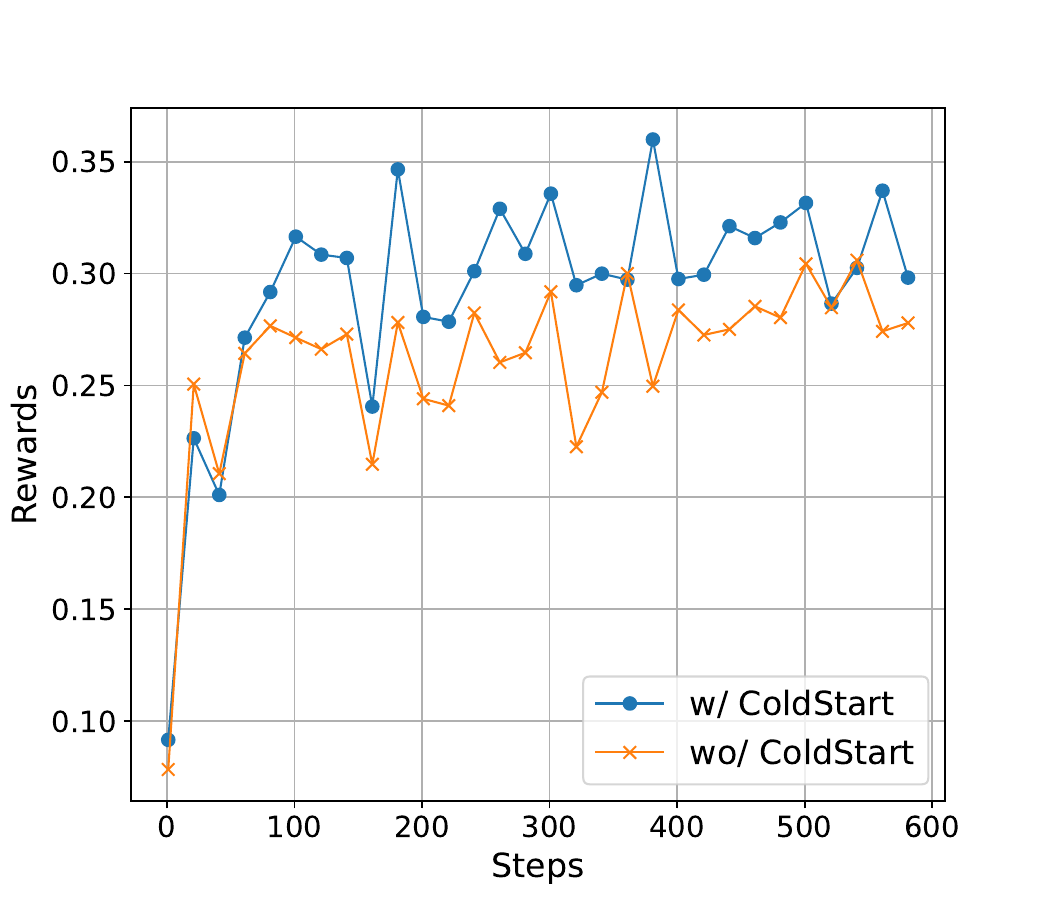}
        \caption{Total Rewards}
    \end{subfigure}%
    \hfill
    \begin{subfigure}[t]{0.31\textwidth}
        \centering
        \includegraphics[width=\textwidth]{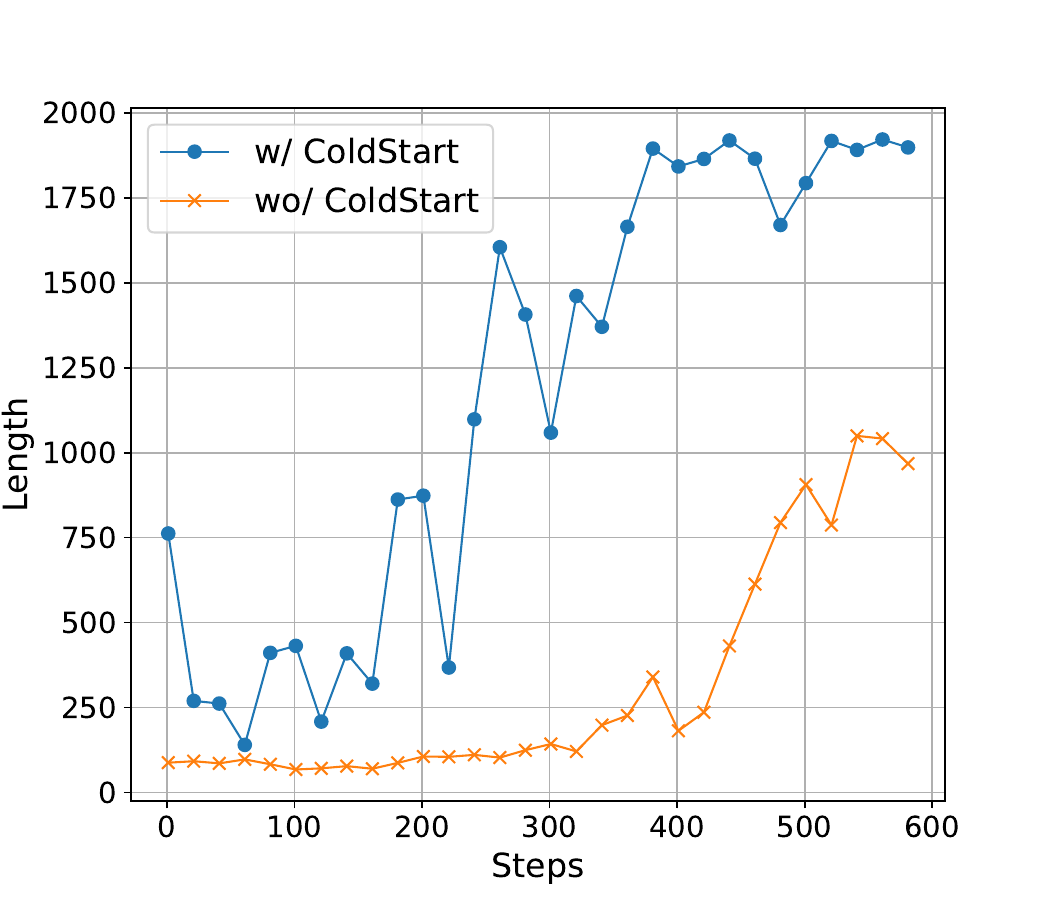}
        \caption{Response Length}
    \end{subfigure}%
    \hfill
    \begin{subfigure}[t]{0.31\textwidth}
        \centering
        \includegraphics[width=\textwidth]{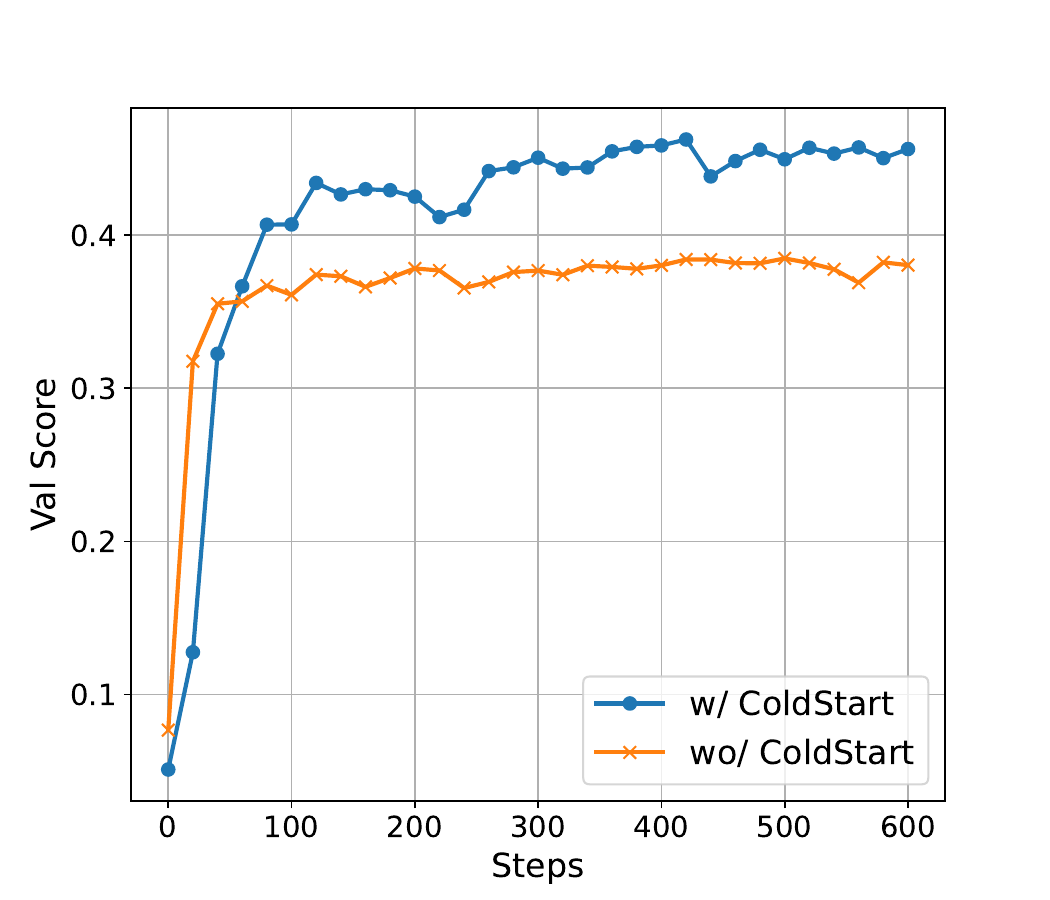}
        \caption{Validation Score}
    \end{subfigure}
    \caption{Comparison of RL training curves with high-quality cold start and without cold start. TVG-R1, with a high-quality cold start, converges to higher scores, demonstrating the benefit of cold start in unlocking the model's potential and enhancing its reasoning abilities, as indicated by the increased response length during training.}
    \label{fig:three-side-by-side}
\end{figure}

\begin{table*}[ht]
\centering
\begin{minipage}{0.49\linewidth}
\centering
\resizebox{\linewidth}{!}{
\begin{tabular}{lcccc}
\toprule
\textbf{Model} & \textbf{Filter} & \textbf{NExTG.} & \textbf{RexT.} & \textbf{Charad.} \\
\midrule
\multicolumn{5}{l}{\textbf{TVG-Coldstart-13k Dataset}} \\
Qwen2.5-VL-7B & - & 20.87 & 8.10 & 46.14 \\
TVG-ColdStart       & -          & 26.14 & 26.26 & 42.19 \\
TVG-R1-U     & $\times$   & 23.92 & \textbf{29.14} & 29.57 \\
TVG-R1              & $\checkmark$ & \textbf{30.41} & 26.38 & \textbf{48.78} \\
TVG-R1-Zero         & -          & 27.76 & 26.00 & 48.75 \\
\bottomrule
\end{tabular}
}
\caption{Validation of the effectiveness of high-quality cold start data. TVG-R1-U refers to performing the cold start on unfiltered data. The results show that TVG-R1 outperforms TVG-R1-U, highlighting the importance of high-quality SFT data.}
\label{tab:coldstart_data}
\end{minipage}
\hfill
\begin{minipage}{0.49\linewidth}
\centering
\resizebox{\linewidth}{!}{
\begin{tabular}{lcccc}
\toprule
\textbf{Model} & \textbf{Filter} & \textbf{NExTG.} & \textbf{RexT.} & \textbf{Charad.} \\
\midrule
\multicolumn{5}{l}{\textbf{TVG-RL-18k Dataset}} \\
Qwen2.5-VL-7B & - &20.87 & 8.10 & 46.14  \\
TVG-R1        & $\times$      & 27.88 & 25.91 & 46.96 \\
TVG-R1        & $\checkmark$  & \textbf{30.41} & \textbf{26.38} & \textbf{48.78} \\
TVG-R1-Zero   & $\times$      & 5.49  & 24.18 & 20.32 \\
TVG-R1-Zero   & $\checkmark$  & 27.76 & 26.00 & 48.75 \\
\bottomrule
\end{tabular}
}
\caption{Validation of the effectiveness of RL data. TVG-R1-Zero refers to skipping the SFT cold start and directly conducting RL training. The results show that RL data filtering improves model performance, particularly in the absence of cold start.}
\label{tab:rl_data}
\end{minipage}
\end{table*}

\section{Experiment}

\subsection{Experimental Setups}

\paragraph{Benchmarks and Evaluation Metrics}
We conduct comprehensive experiments on three benchmarks to evaluate the effectiveness of our approach. Specifically, we report results on the ReXTime~\cite{rextime}, NExT-GQA~\cite{nextgqa}, and Charades-STA~\cite{charadessta} datasets.
For evaluation, we adopt the R1@m metric for temporal video grounding (TVG). R1@m denotes the percentage of instances where the top-1 predicted segment achieves an Intersection-over-Union (IoU) greater than a threshold $m$, where $m \in {0.3, 0.5, 0.7}$. Additionally, we report the mean IoU (mIoU) across all samples as an overall indicator of TVG accuracy.

\paragraph{Baselines}
We compare our approach with several strong baselines, including instruction-tuned temporal localization models such as VTimeLLM~\cite{Huang2024VTimeLLM}, TimeChat~\cite{Ren2024TimeChat}, and VideoChat-TPO~\cite{yan2025task}, as well as general-purpose multimodal large models like Qwen2.5-VL 7B and 32B~\cite{bai2025qwen2}. For models marked with ``thinking,'' we employ the TVG-R1 prompt template to guide temporal grounding.



\paragraph{Training Details.}
All experiments are conducted on 16 NVIDIA H100 (80GB) GPUs. For both training and inference, we limit the number of video frames to 64, with each frame processed at a resolution of $128 \times 28 \times 28$ pixels. The backbone model is Qwen2.5-VL-7B~\cite{bai2025qwen2}. The hyperparameters $\epsilon_1$ and $\epsilon_2$ are set to 0.8 and 0.4, respectively.
We first perform supervised fine-tuning (SFT) on the TVG-Coldstart-13K dataset for one epoch to obtain the TVG-cold start model. Next, we apply reinforcement learning (RL) on the TVG-RL-18K dataset to obtain the final TVG-R1 model, where the hyperparameter $\beta$ in the KL divergence term of the GRPO algorithm is set to 0.0. The maximum response length is set to 2048 tokens, and the loss weights $\lambda_{\text{tIoU}}$ and $\lambda_{\text{form}}$ are set to 0.9 and 0.1, respectively. Due to computational resource constraints, RL training is limited to 600 steps.
Additional details can be found in the Appendix.

\begin{table}[htbp]
\centering
\resizebox{0.5\linewidth}{!}{
\begin{tabular}{lcccc}
\toprule
\textbf{Model} & \textbf{R@0.3} & \textbf{R@0.5} & \textbf{R@0.7} & \textbf{mIoU} \\
\midrule
Base              & 68.98 & 48.18 & 22.87 & 46.14  \\
Base thinking     & 36.48 & 21.83 & 9.76     & 23.48 \\
\midrule
VTimeLLM          & 55.3  & 34.3  & 14.7  & 34.6  \\
TimeChat          & 51.5  & 32.2  & 13.4  & -     \\
VideoChat-TPO     & 58.3  & 40.2  & 18.4  & 38.1  \\
\midrule
TVG-ColdStart & 42.23 & 29.38 & 14.95 & 28.91    \\
TVG-R1            & \textbf{70.75} & \textbf{50.46} & \textbf{23.92}    & \textbf{46.73} \\
\bottomrule
\end{tabular}
}
\caption{Performance comparison on Charades dataset}
\label{tab:charades_main_results}
\end{table}

\subsection{Main Results}

As shown in Tables~\ref{tab:main_results} and \ref{tab:charades_main_results}, our experiments across three different benchmarks demonstrate the effectiveness of VTG-R1 on video temporal grounding tasks. Two key observations can be drawn.

\textbf{Outstanding Performance of VTG-R1:} VTG-R1 consistently outperforms previous models on most benchmarks, highlighting the importance of explicit reasoning in addressing video temporal grounding challenges. These results further underscore the impact of reinforcement learning in boosting model performance.

\textbf{Importance of Reinforcement Learning:} The SFT-based model, TVG-ColdStart, does not consistently yield performance gains and even exhibits a slight decrease after SFT, possibly due to overfitting or limited generalization to unseen scenarios. In contrast, after reinforcement learning, VTG-R1 achieves substantial improvements, strongly suggesting that RL is essential for developing robust reasoning capabilities that generalize effectively.

\subsection{Analysis}

To gain deeper insights into the impact of different variants, we present experimental results under additional configurations. Specifically, we analyze variants associated with cold start process and RL data selection.

\paragraph{Finding 1: High-Quality cold start data is crucial.}

As shown in Fig.~\ref{fig:three-side-by-side}, we compare the RL training curves of TVG-R1 and TVG-R1-Zero, where TVG-R1-Zero refers to skipping the SFT cold start and directly performing RL training. It can be observed that, in terms of both total rewards during training and test set performance, TVG-R1 converges to higher scores, suggesting that a high-quality cold start helps unlock the model's potential in the RL phase. Furthermore, as illustrated in Fig.~\ref{fig:three-side-by-side}(b), the model initialized with a cold start exhibits a higher response length at the outset, with a more pronounced increase throughout training. This indicates that the cold start enhances the model's reasoning ability, enabling it to derive correct answers through more detailed reasoning.

We further examine the impact of cold start response length on model performance by limiting the maximum output length of Gemini-2.5-Pro. We re-annotate different cold start datasets, and the final results after RL training are reported in Table~\ref{tab:max-length-settings}. The results indicate that longer response lengths during the cold start phase are more beneficial for model optimization.

Additionally, as shown in Table~\ref{tab:coldstart_data}, we compare TVG-R1 and TVG-R1-U, where TVG-R1-U denotes using the unfiltered 56K dataset for cold start followed by RL. Note that all RL is performed on the TVG-RL-18K dataset. The results show that TVG-R1 significantly outperforms TVG-R1-U, demonstrating that selecting high-quality cold start data is more effective for learning robust reasoning abilities than simply increasing the quantity of training data.


\begin{table}[tbp]
\centering
\resizebox{0.5\linewidth}{!}{
\begin{tabular}{cccc}
\toprule
\textbf{Max Length} & \textbf{NExTGQA} & \textbf{RexTime} & \textbf{Charades} \\
\midrule
2048 & 30.41 & 26.38 & 48.78 \\
1024 & 21.80 & 25.71 & 41.38 \\
512  & 24.09 & 24.91 & 46.31 \\
\bottomrule
\end{tabular}
}
\caption{Impact of cold start length on performance. The results after RL training show that longer response lengths during cold start are more beneficial for the model's optimization.}
\label{tab:max-length-settings}
\end{table}

\paragraph{Finding 2: Controlling the difficulty of RL training data is necessary.}

As shown in Table~\ref{tab:rl_data}, we compare the results of RL training with and without data filtering under both the TVG-R1 and TVG-R1-Zero settings. Note that TVG-R1 is initialized with the TVG-Coldstart-13K dataset. The results indicate that, without cold start, models trained on unfiltered data struggle to learn, whereas data filtering leads to substantial performance improvements. Moreover, for models initialized with cold start, filtering the RL data further benefits model optimization. These findings suggest that if the training data is too challenging or confusing in the early stages, the model may have difficulty learning and achieving convergence. 



\section{Conclusion}
In this work, we present a novel two-stage training framework for Video Temporal Grounding (VTG) to enhance the capabilities of large vision-language models. Extensive experiments on multiple benchmarks demonstrate that high-quality cold-start data and difficulty-controlled RL training are both crucial for improving model performance and generalization. Our approach is shown to be scalable and effective for real-world deployment. 

\section*{Limitations}

While our proposed framework demonstrates significant improvements for Video Temporal Grounding (VTG), several limitations remain. First, the approach relies heavily on high-quality, curated cold-start data, which may be difficult to obtain in certain domains or low-resource scenarios. Second, the reinforcement learning stage introduces considerable computational overhead, potentially limiting accessibility for smaller organizations or academic users with constrained resources.  Future work should explore ways to improve data efficiency, optimize RL for resource-limited settings, and broaden the applicability of this training paradigm to more complex or diverse multimodal tasks.


\newpage

\bibliographystyle{plainnat}
\bibliography{main}

\newpage
\appendix

\section{Implementation Details}
\label{sec:appendix}

\subsection{Recipes}
\paragraph{TVG-Coldstart Dataset}
We use gemini-2.5-pro-preview-05-06 API for annotation and set the max length to 8192. Samples with IoU larger than 0.8 are selected for coldstart.

\paragraph{Coldstart Stage}
We finetune the base model on the TVG-Coldstart dataset. The finetuning is performed on 8 H100 GPUs with batch size 8 for 1 epoch. The learning rate is set to 1e-6.

\paragraph{RL Stage}
We perform RL training base on Easy-R1~\cite{zheng2025easyr1} implementations. The maximum response length is set to 2048. The batch size is set to 128 and trained for 600 steps. The number of GRPO samples $G$ is set to 8.

\subsection{Experiments}
Evaluations are conducted using the official VideoMind~\cite{liu2025videomind} implementation. The maximum response length is set to 2048 tokens, and all other inference hyperparameters are kept at their default values as provided by the \texttt{transformers} library.

\section{Qualitative Result}

\subsection{TVG-R1 Evaluation Cases}
\begin{figure*}[htbp]
    \centering
    \includegraphics[width=1.0\textwidth]{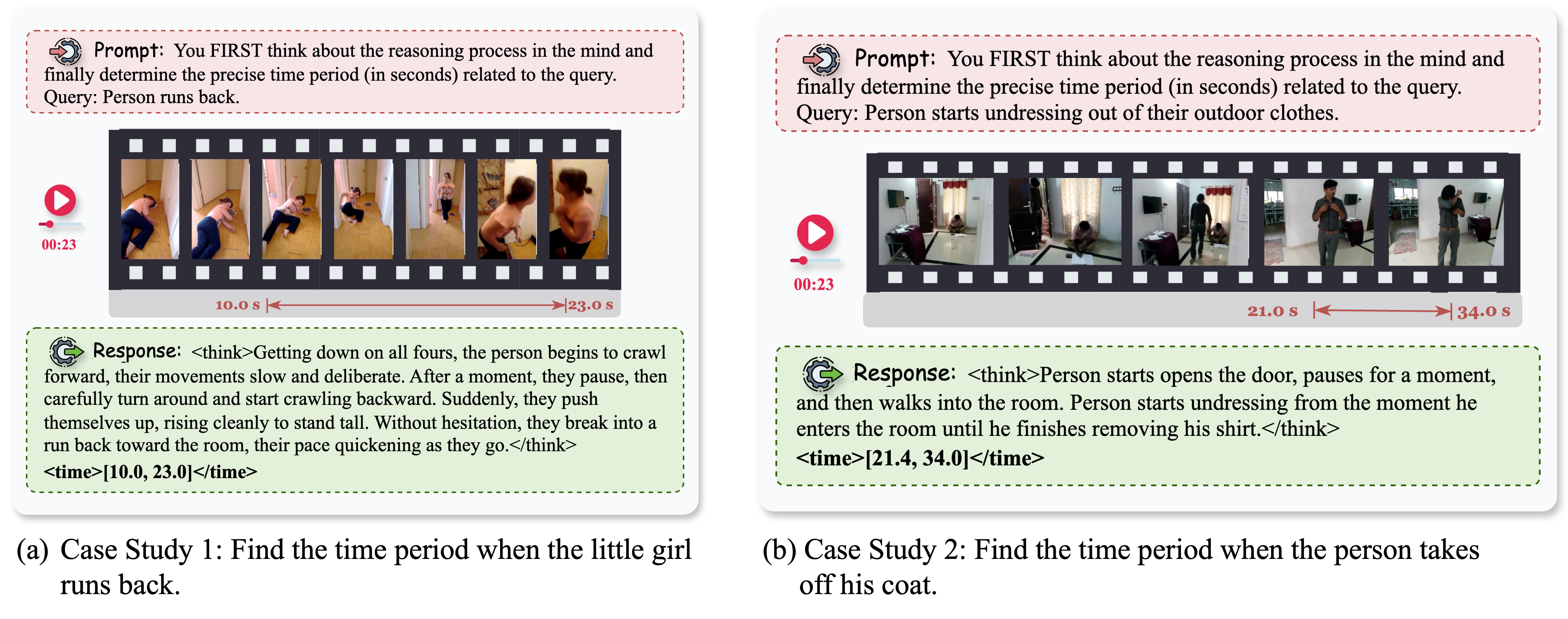}
    \caption{\textbf{TVG-R1 Evaluation Cases}.}
    \label{fig:evaluation_case_study}
\end{figure*}

We provide qualitative cases for TVG-R1 in Fig.~\ref{fig:evaluation_case_study}.
These comprehensive data document the reasoning process and prediction results of temporal localization models in video segment understanding and localization tasks. Each data entry includes basic video information, the query, the annotated time span, the model’s step-by-step reasoning process, and the predicted time span. The reasoning content typically provides a detailed description of the sequence of key events and action nodes in the video, helping the model clarify the start and end points of the target segment. The prediction results are evaluated by comparing them with the ground-truth spans using metrics such as IoU. This type of data not only highlights the combination of multi-step reasoning and temporal cues, but also reflects the model’s localization capability in concrete cases, providing a solid basis for performance evaluation and analysis.

\subsection{TVG-Coldstart Dataset Cases}

\begin{figure*}[htbp]
    \centering
    \includegraphics[width=1.0\textwidth]{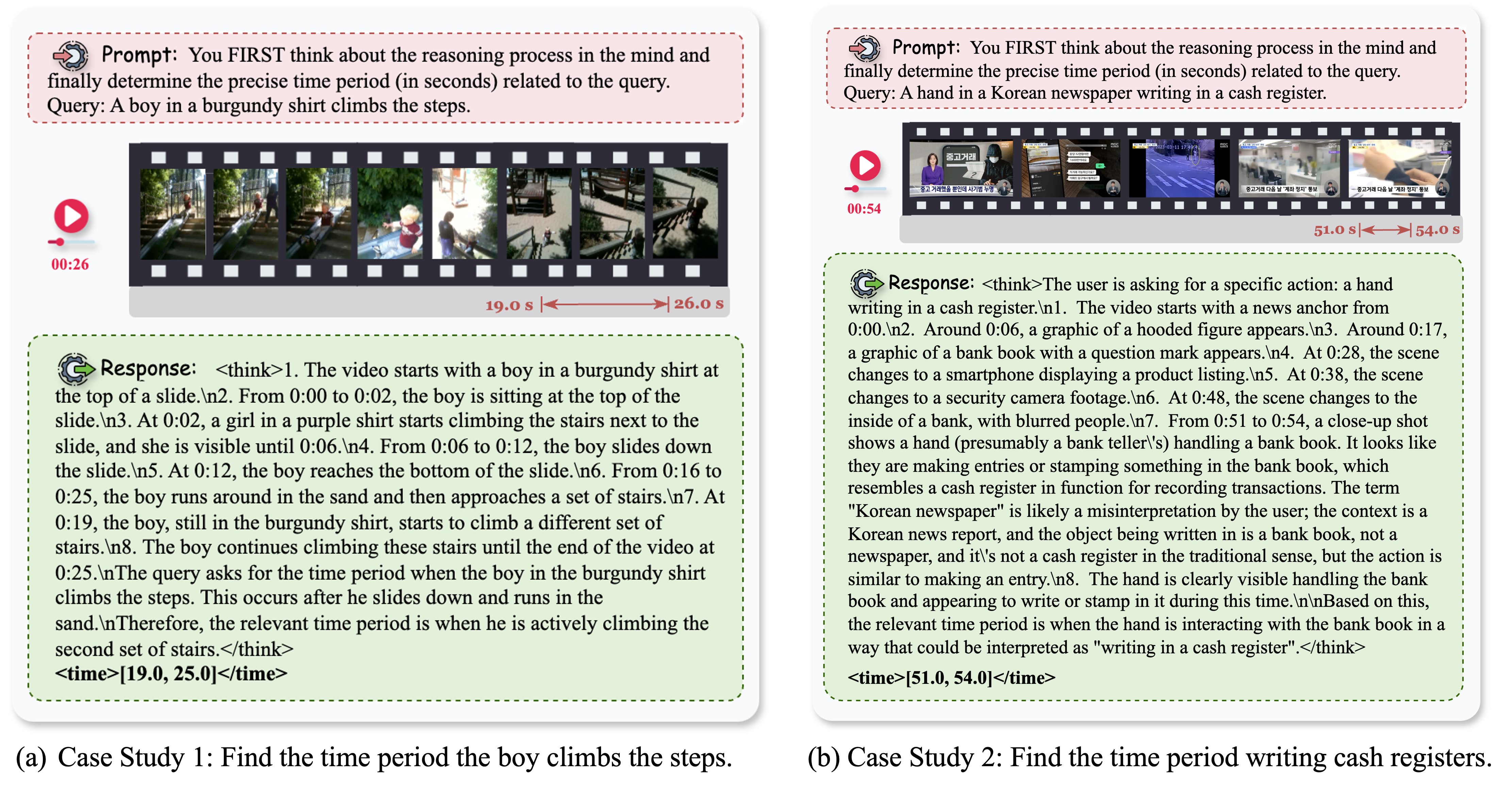}
    \caption{\textbf{TVG-Coldstart Dataset Cases}.}
    \label{fig:cold_start_case_study}
\end{figure*}

We provide qualitative cases for TVG-Coldstart Dataset in Fig.~\ref{fig:cold_start_case_study}. These cold-start data samples employ step-by-step reasoning to document the model’s decision-making process for temporal localization tasks. The data cover the identification of key actions and event nodes within video segments, clearly illustrating how the model analyzes each segment and filters events to pinpoint the exact time period required by the query. This type of data emphasizes multi-step reasoning combined with temporal cues, providing high-quality reasoning samples for the subsequent training and evaluation of video understanding models.

\section{Additional Related Works}

Early studies indicate that Reinforcement Learning with Human Feedback (RLHF) can effectively align Large Language Models (LLMs) with human preferences, primarily ensuring large models to follow human intentions and values~\cite{stiennon2020learning, bai2022training, ouyang2022training, chen2025diffpo, chen2023fast, chen2024identifying, chen2024pad, chen2024learnable, fan2024fairmt, fan2024biasalert}. More recent research has shifted attention toward Reinforcement Learning with Verifiable Reward (RLVR) for tasks characterized by deterministic answers~\cite{yu2025dapo, liu2025understanding, feng2025mt}. 

As a pioneering open-source LLM, DeepSeek-R1~\cite{guo2025deepseek} employs Generative Reward-driven Policy Optimization (GRPO)~\cite{shao2024deepseekmath} to augment its reasoning performance, leveraging carefully designed rule-based rewards that integrate both reasoning templates and final outcomes. Within the context of LVLMs, recent methodologies have applied GRPO to multimodal image reasoning tasks, thereby substantially improving image comprehension~\cite{meng2025mm,huang2025vision} and video understanding~\cite{wang2025time,feng2025video}.

\end{document}